\documentclass[runningheads]{llncs}
\usepackage[T1]{fontenc}
\usepackage{graphicx}
\usepackage[utf8]{inputenc}
\usepackage{mathrsfs}
\usepackage{scalerel}
\usepackage{url}
\usepackage{bbold}
\usepackage{amsfonts}
\usepackage{amsmath,amssymb,amsfonts}
\usepackage{wrapfig}
\usepackage{mathrsfs} 
\usepackage{algorithm,algorithmic}
\usepackage{array}
\usepackage{times}
\usepackage{url}
\usepackage{subfigure}
\usepackage{cite}
\usepackage{upgreek}
\usepackage{float}
\usepackage{longtable}
\usepackage{color}
\usepackage{wasysym}
\usepackage{grffile}
\newcommand{\1}{\mathbb{1}}

\newcommand{\E}{\mathbb{E}}
\newcommand{\R}{\mathbb{R}}

\newcommand{\Db}{\mathbf{D}}

\newcommand{\alphab}{\boldsymbol{\alpha}}

\newcommand{\Phib}{\boldsymbol{\Phi}}

\newcommand{\thetab}{\boldsymbol{\theta}}

\newcommand{\Lambdab}{\boldsymbol{\Lambda}}

\newcommand{\Dc}{\mathcal{D}}

\newcommand{\Xc}{\mathcal{X}}

\begin{document}
%
\title{Tracking Dynamic Gaussian Density with a Theoretically Optimal Sliding Window Approach}

%
%
\author{Yinsong Wang\inst{1}\orcidID{0000-0001-8232-5586} \and
Yu Ding\inst{2}\orcidID{0000-0001-6936-074X} \and
Shahin Shahrampour\inst{1}\orcidID{0000-0003-3093-8510}}
\authorrunning{Y. Wang et al.}
%
\institute{Northeastern University, Boston MA 02115, USA \and
Georgia Institute of Technology, Atlanta GA 30332, USA}
\maketitle              
\begin{abstract}

Dynamic density estimation is ubiquitous in many applications, including computer vision and signal processing. One popular method to tackle this problem is the "sliding window" kernel density estimator. There exist various implementations of this method that use  heuristically defined weight sequences for the observed data. The weight sequence, however, is a key aspect of the estimator affecting the tracking performance significantly. In this work, we study the exact mean integrated squared error (MISE) of "sliding window" Gaussian Kernel Density Estimators for evolving Gaussian densities. We provide a principled guide for choosing the optimal weight sequence by theoretically characterizing the exact MISE, which can be formulated as constrained quadratic programming. We present empirical evidence with synthetic datasets to show that our weighting scheme indeed improves the tracking performance compared to heuristic approaches.

\end{abstract}
\section{Introduction}
Dynamic density estimation is an important topic in various applications, such as manufacturing, sensor networks, and traffic control. One popular choice for tackling this problem is the "sliding window" kernel density estimator \cite{zhou2003m, heinz2008cluster, qahtan2016kde, wang2022takde}. This class of estimators has shown impressive performance in practice, and continuous studies on improving the method have contributed to its success. However, to the best of our knowledge, most existing works focus on the kernel function itself. For example, M-kernel method \cite{zhou2003m} merges data points to the closest previously defined grid points. Cluster kernel and resampling techniques \cite{heinz2008cluster} further improve the merging performance, assuming exponentially decaying importance for older data points. Local region kernel density estimator \cite{boedihardjo2008framework} varies the kernel bandwidth across different regions in the density support. Adaptive bandwidth method \cite{amiri2018density} updates the kernel bandwidth sequence as new data points are observed. All these studies put a heavy emphasis on the kernel function but use heuristic approaches in weighting the observed data points. 

A recent development in natural language processing introduced the attention mechanism \cite{vaswani2017attention} for sequential data modeling, which has seen immense success in various fields. The method highlights the importance of correlations between sequential data points, which can be effectively modeled with a weight sequence that captures the importance of each data point. For example, the state-of-the-art natural language processing model BERT \cite{devlin2018bert} represents each word as a weighted average of other words in a sentence to make predictions. Graph attention networks \cite{velivckovic2017graph} represent each node feature as a weighted average of other nodes features to capture structural information. These developments motivate us to revisit "sliding window" kernel density estimators and improve upon heuristic weighting schemes.

In this work, we investigate the theoretical aspect of dynamic Gaussian density tracking using a "sliding window" Gaussian kernel density estimator. We calculate the exact estimation accuracy in terms of mean integrated squared error (MISE) and show the impact of the weight sequence on the MISE. We prove that MISE can be formulated as a constrained quadratic programming that can lead us to a unique optimal weight sequence. We provide numerical experiments using synthetic dynamic Gaussian datasets to support our theoretical claim that this weighting scheme indeed improves the tracking performance compared to heuristic approaches.

\section{Problem Formulation and Algorithm}


We focus on dynamic data arriving in batches, where at each time $t$ we observe a batch of $n_t$ data points $\{x_j^{(t)}\}_{j=1}^{n_t}$. This data structure applies to many real-world time-series datasets \cite{UCRArchive2018}. We assume that at time $t$, the data points are sampled from a Gaussian distribution, i.e., the true density has the following form
\begin{equation*}
    p_t(x) = \phi_{\gamma_t}(x-\mu_t) \triangleq \frac{1}{\sqrt{2\pi}\gamma_t}e^{\frac{-(x-\mu_t)^2}{2\gamma_t^2}}.
\end{equation*}
The density evolution can then be uniquely identified with sequences of means and standard deviations. For a dataset $\Dc$ over a time span of $m$, we have $m$ batches, i.e., $\Dc = \{\Xc_i\}_{i=1}^m$, where $\Xc_i = \{x_j^{(i)} \sim N(\mu_i, \gamma_i^2)\}_{j=1}^{n_i}$ and
$N(\mu_i, \gamma_i^2)$ denotes the Gaussian distribution with mean $\mu_i$ and standard deviation $\gamma_i$.

\vspace{.2cm}
\noindent
{\bf "Sliding Window" Kernel Density Estimator:} Let us restrict our attention to Gaussian kernel density estimators where
\begin{equation*}
    K_{\sigma}(x - x') = \phi_{\sigma}(x-x') \triangleq \frac{1}{\sqrt{2\pi}\sigma}e^{\frac{-(x-x')^2}{2\sigma^2}},
\end{equation*}
where $\sigma$ denotes the kernel bandwidth. Then, the dynamic density estimator has the following form
\begin{equation}\label{DKDE}
    \hat{h}_t(x) = \sum_{i=1}^T \alpha_i\hat{p}_i(x)=\sum_{i=1}^T \frac{\alpha_i}{n_i}\sum_{j=1}^{n_i}K_{\sigma}(x-x^{(i)}_j),
\end{equation}
where a window size of $T$ previous batches is taken into account. The weights $\alphab^{(t)}\triangleq[\alpha_1^{(t)},\ldots,\alpha_T^{(t)}]^\top$ vary over time $t$, but the superscript $(t)$ is omitted for the presentation simplicity. Each "sliding window" kernel density estimator is essentially a weighted average of kernel density estimators, and using $\sum_{i=1}^T \alpha_i = 1$, we can ensure that the estimator \eqref{DKDE} is a proper density function.

\vspace{.2cm}
\noindent
{\bf Mean Integrated Squared Error:} MISE is a popular metric to characterize the accuracy of density estimators \cite{marron1992exact, wand1994kernel}. For any density estimator $\hat{h}(x)$ and the true density $p(x)$, the MISE is formally defined as
\begin{equation*}
    MISE(\hat{h}(x)) \triangleq \int \E[(\hat{h}(x) - p(x))^2] dx,
\end{equation*}
where the expectation is taken over the randomness of the data samples (sampled from density $p(x)$), and the integral over $x$ accounts for the accumulated error over the support of the density function. MISE is often decomposed into {\it bias} and {\it variance} terms as follows
\begin{equation}\label{eq:MSEdecomp}
    \begin{aligned}
    MISE(\hat{h}(x)) 
    & = \int \E[(\hat{h}(x)-\E[\hat{h}(x)])^2]+(\E[\hat{h}(x)]-p(x))^2 dx\\
    &= \int V(\hat{h}(x)) + B^2(\hat{h}(x)) dx= IV(\hat{h}(x)) + IB^2(\hat{h}(x)),
    \end{aligned}
\end{equation}
where $V(\cdot)$ and $B^2(\cdot)$ are called the variance and squared bias, and $IV(\cdot)$ and $IB^2(\cdot)$ are called the integrated variance and integrated squared bias, respectively.

\section{Theoretical Result: Optimal Weight Sequence}
We now present our main theorem, which states that the exact MISE of the "sliding window" kernel density estimators of evolving Gaussian densities can be calculated, and it is a quadratic function of the weight sequence $\alphab$.
\begin{theorem}\label{the: mise}
Estimating the evolving Gaussian density $p_t(x)$ with the "sliding window" Gaussian Kernel Density Estimator \eqref{DKDE} results in the following exact mean integrated squared error
\begin{equation*}
    MISE(\hat{h}_t(x)) = \alphab^\top \Lambdab \alphab - 2\thetab^\top\alphab + \frac{1}{2\gamma_t\sqrt{\pi}},
\end{equation*}
where $\Lambdab = \Phib + \Db$, and $\Phib \in \R^{T \times T}$ is such that $[\Phib]_{ij}=\phi_{(\sigma^2+\gamma^2_i+\sigma^2+\gamma^2_j)^{1/2}}(\mu_i-\mu_j)$. $\Db  \in \R^{T \times T}$ is a diagonal matrix with $[\Db]_{ii}=\frac{1}{n_i2\sqrt{\pi}}(\frac{1}{\sigma} - \frac{1}{\sqrt{\sigma^2+\gamma^2_i}})$, and the $i$-th entry of vector $\thetab$ is $\phi_{(\sigma^2+\gamma^2_i+\gamma_t^2)^{1/2}}(\mu_i-\mu_t)$. To connect with \eqref{eq:MSEdecomp}, we have the following
\begin{equation*}
    \begin{aligned}
        IB^2(\hat{h}(x)) &= \alphab^\top \Phib \alphab - 2\thetab^\top\alphab + \frac{1}{2\gamma_t\sqrt{\pi}},\\
        IV(\hat{h}(x)) &= \alphab^\top \Db \alphab.
    \end{aligned}
\end{equation*}
\end{theorem}
The proof of Theorem \ref{the: mise} can be found in the Appendix. We are now able to propose the following corollary, which suggests that one can find the optimal weight sequence for the estimator \eqref{DKDE} by optimizing MISE over weights.   
\begin{corollary}\label{cor:weight}
The optimal weight sequence under MISE for the dynamic density estimation is determined by the following constrained quadratic programming 
    \begin{equation*}
    \begin{aligned}
        \underset{\alphab}{min} \quad & \alphab^\top \Lambdab \alphab - 2\thetab^\top\alphab + \frac{1}{2\gamma_t\sqrt{\pi}}\\
        s.t. \quad & \1^\top\alphab = 1,\quad \alpha_i \geq 0, \\
    \end{aligned}
\end{equation*}
where $\1$ is the vector of all ones.
\end{corollary}

\section{Empirical Results}
In this section, we investigate the empirical behavior of the dynamic kernel density estimator for different weighting methods. We first design synthetic dynamic datasets with evolving Gaussian densities following the foundation of our theoretical claim. Then, we measure the MISE of the density estimation over different experimental settings to validate our theoretical results.

\subsection{Synthetic Dataset Design}

We consider a synthetic dynamic dataset for a time span of $100$ batches. Each batch of data points is sampled from an evolving Gaussian density. The design principles of the synthetic dataset are as follows:
\begin{itemize}
    \item The evolution in the mean of the Gaussian distribution follows a random walk model with the starting point of $0$, i.e.,
    \begin{equation*}
        \mu_{t+1} = \mu_t + \epsilon^{(t)}_{1}, 
    \end{equation*}
    where $\mu_0 = 0$, and $\epsilon^{(t)}_1$ denotes a sample from the uniform distribution on $[-1,1]$.
    \item The evolution in the standard deviation of the Gaussian distribution follows a lower bounded random walk model with the starting point of 1, i.e.,
    \begin{equation*}
        \gamma_{t+1} = \max \{\gamma_t + \epsilon^{(t)}_{0.2}, \gamma_0\}, 
    \end{equation*}
    where $\gamma_0 = 1$, and $\epsilon^{(t)}_{0.2}$ denotes a sample from the uniform distribution on $[-0.2,0.2]$.
    \item Each batch of data randomly contains $3$ to $20$ data points, which limits the amount of available data  at each time to justify the use of dynamic density estimators and also highlights the impact of data availability in the weightings.
\end{itemize}

\subsection{Experimental Settings}

In our experiment, we compare the optimal weighting sequence proposed in Corollary \ref{cor:weight} (denoted by "dynamic") with three other popular baseline methods. 
\begin{itemize}
    \item {\bf Current:} The traditional kernel density estimator (KDE), which only uses data points from the current batch for estimation. Basically, the last element of the weight sequence $\alphab \in \R^T$ is equal to $1$, and all other elements are $0$.
    \item {\bf Average:} This method combines all data points in the window by equally weighting all batches. That is to say, all elements of $\alphab \in \R^T$ are equal to $1/T$.
    \item {\bf Exponential:} This is one of the most popular weighting methods in common practice. It assumes the correlations between data points decay exponentially over time. The weight sequence $\alphab \in \R^T$ with a decay factor $\beta$ is as follows
    \begin{equation*}
        \alpha_1 = (1-\beta)^{T-1}, \quad \{\alpha_i = (1-\beta)^{T-i}\beta\}_{i=2}^T.
    \end{equation*}
    In our simulation, we set $\beta = 0.1$, which shows the best performance for this weighting method.
\end{itemize}
We perform the comparison over two variables, namely the window size $T$ and the kernel bandwidth $\sigma$. When comparing different window sizes, we fix the kernel bandwidth for all estimators as $\sigma = 1$, which is a good kernel bandwidth choice for the baseline weighting methods according to our bandwidth comparison. When comparing different bandwidths, we fix the window size $T=5$, which is also in favor of the baseline weighting methods according to our simulation. 

We carry out $20$ Monte-Carlo simulations to generate error bars, where we randomly create a brand new synthetic dataset for each Monte-Carlo simulation.

\subsection{Performance}
\begin{figure}[t]
\includegraphics[width=0.45\textwidth]{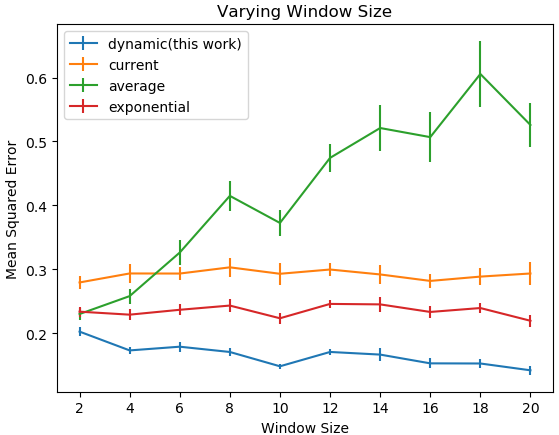}
\includegraphics[width=0.45\textwidth]{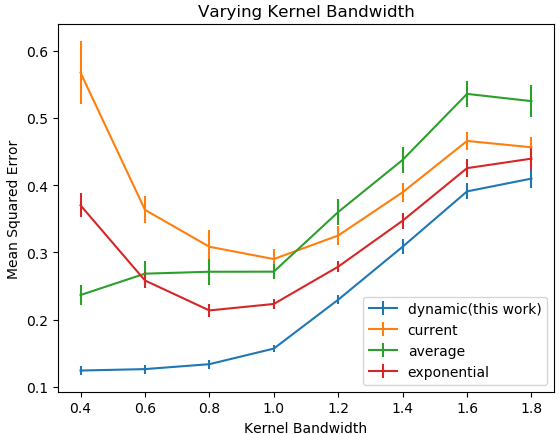}
\caption{{\bf Left:} The estimated MISE versus different window sizes. {\bf Right:} The estimated MISE versus different kernel bandwidths.} \label{fig: error}
\end{figure}
The estimated MISE for all methods is shown in Fig. \ref{fig: error}. We observe that the optimal weighting sequence (dynamic) achieves the best performance (lowest MISE) in all cases. 

For the window size comparison, the current weighting KDE serves as a baseline, where the error does not (statistically) change with different window sizes. We see that the average weighting has a significant error as window size increases. The exponential weighting is consistently better than the current weighting. However, the dynamic weighting achieves the best performance for all window sizes. This observation, together with our theoretical characterization, shed light on why the "sliding window" kernel density estimator works better than the traditional KDE for dynamic density estimation. It introduces a mild bias from previous data points in favor of reducing the variance (induced by low data volume) to improve the estimation accuracy.

For the kernel bandwidth comparison, the current weighting KDE again serves as a baseline, where the optimal performance occurs at $\sigma = 1$. We can see that (for $\sigma\leq 1$) all "sliding window" estimators are better than KDE under a relatively small window size of $5$ (low bias), and the optimal weighting method (dynamic) is consistently better than all other benchmark methods.

\section{Conclusion}

In this work, we theoretically characterized the MISE of the "sliding window" kernel density estimator for evolving Gaussian densities as a quadratic function of the weight sequence. Our result underscores the important role of the weight sequence in the estimation accuracy. We also provided numerical experiments to support our theoretical claim. For future directions, we would like to expand this theory to Gaussian mixtures, which can potentially apply to any dynamic density tracking problem.

\section{Appendix: Proof of Theorem \ref{the: mise} }
We first state a (standard) property of Gaussian densities in the following lemma.

\begin{lemma}\label{LE: normalproduct}
For Gaussian density functions the following relationship holds
\begin{equation*}
    \phi_a(x-\mu_a)\phi_b(x-\mu_b) = \phi_{(a^2+b^2)^{1/2}}(\mu_a-\mu_b)\phi_{\frac{ab}{(a^2+b^2)^{1/2}}}(x-\mu_{ab}),
\end{equation*}
where $\mu_{ab} = \frac{a^2\mu_b+b^2\mu_a}{a^2+b^2}$.
\end{lemma}

Using \eqref{DKDE}, we can write the bias of the estimator in \eqref{eq:MSEdecomp} as
\begin{equation}
\begin{aligned}\label{bias}
    B(\hat{h}_t(x)) &\triangleq \E[\hat{h}_t(x)-p_t(x)]=\E\Big[\sum_{i=1}^T \frac{\alpha_i}{n_i}\sum_{j=1}^{n_i}K_{\sigma}(x-x^{(i)}_j)-p_t(x)\Big]\\
    &= \sum_{i=1}^T\alpha_i\int K_{\sigma}(x-y)p_i(y)dy -p_t(x)=\sum_{i=1}^T\alpha_i(K_{\sigma}*p_i)(x)-p_t(x),
\end{aligned}
\end{equation}
where $*$ denotes the convolution, and $p_i(\cdot)$ is the true density of batch $i$. The estimator variance in \eqref{eq:MSEdecomp} can also be calculated as
\begin{equation*}
\begin{aligned}
V(\hat{h}_t(x)) = \sum_{i=1}^T \alpha^2_iV(\hat{p}_i(x)),
    \end{aligned}
\end{equation*}
due to the independence, where
\begin{equation}\label{variance}
    V(\hat{p}_i(x)) = \frac{1}{n_i}\Big((K^2_{\sigma}*p_i)(x)-(K_{\sigma}*p_i)^2(x)\Big).
\end{equation}
Given the expressions of bias \eqref{bias} and variance \eqref{variance}, to calculate the exact MISE, we need to characterize the quantities $(K^2_{\sigma}*p_i)(x)$ and $(K_{\sigma}*p_i)(x)$. First, we consider $(K_{\sigma}*p_i)(x)$ and use Lemma \ref{LE: normalproduct} to get
\begin{equation*}
    \begin{aligned}
        (K_{\sigma}*p_i)(x) & = \int \phi_{\sigma}(x-y)\phi_{\gamma_i}(y-\mu_i) dy
        = \phi_{(\sigma^2+\gamma^2_i)^{1/2}}(x-\mu_i).
    \end{aligned}
\end{equation*}
We further characterize $(K^2_{\sigma}*p_i)(x)$ as following
\begin{equation*}
    \begin{aligned}
        (K^2_{\sigma}*p_i)(x) &= \int \phi^2_{\sigma}(x-y)\phi_{\gamma_i}(y-\mu_i) dy\\
        &= \frac{1}{2\sigma\sqrt{\pi}} \int \phi_{\sigma/\sqrt{2}}(x-y)\phi_{\gamma_i}(y-\mu_i) dy\\
        &= \frac{1}{2\sigma\sqrt{\pi}} \phi_{(\sigma^2/2+\gamma^2_i)^{1/2}}(x-\mu_i).
    \end{aligned}
\end{equation*}
Having the above expressions, we can calculate the integrated variance term as
\begin{equation*}
\begin{aligned}
    IV(\hat{p}_i(x)) 
    =\frac{1}{n_i2\sqrt{\pi}}\bigg(\frac{1}{\sigma} - \frac{1}{\sqrt{\sigma^2+\gamma^2_i}}\bigg),
\end{aligned}
\end{equation*}
using Lemma \ref{LE: normalproduct}. The exact MISE depends also on the integrated bias square \eqref{eq:MSEdecomp}, which takes the following expression
\begin{equation*}
    \begin{aligned}
    IB(\hat{h}_t(x))^2 = \int \bigg(\sum_{i=1}^T\alpha_i\phi_{(\sigma^2+\gamma^2_i)^{1/2}}(x-\mu_i) - \phi_{\gamma_t}(x-\mu_t)\bigg)^2 dx.
    \end{aligned}
\end{equation*}
There exist three types of terms in the above expression, which we examine one by one. First, we look at the interaction terms of the following form
\begin{equation*}
\resizebox{0.99\hsize}{!}{$ 
    \int \alpha_i\phi_{(\sigma^2+\gamma^2_i)^{1/2}}(x-\mu_i)\alpha_j\phi_{(\sigma^2+\gamma^2_j)^{1/2}}(x-\mu_j)dx=\alpha_i\alpha_j\phi_{(\sigma^2+\gamma^2_i+\sigma^2+\gamma^2_j)^{1/2}}(\mu_i-\mu_j),$}
\end{equation*}
which follows from Lemma \ref{LE: normalproduct}. The second type is the interactions terms where
\begin{equation*}
    \begin{aligned}
    \int \alpha_i\phi_{(\sigma^2+\gamma^2_i)^{1/2}}(x-\mu_i)\phi_{\gamma_t}(x-\mu_t) dx = \alpha_i\phi_{(\sigma^2+\gamma^2_i+\gamma_t^2)^{1/2}}(\mu_i-\mu_t).
    \end{aligned}
\end{equation*}
The last term is the square of $p_t(x)$, for which we have $\int p^2_t(x) dx = \frac{1}{2\gamma_t\sqrt{\pi}}$.

Given the above expressions, we can write the square integrated bias as a quadratic function of weight sequence $\alphab$ as follows
\begin{equation*}
    IB(\hat{h}_t(x))^2 = \alphab^\top \Phib \alphab - 2\thetab^\top\alphab + \frac{1}{2\gamma_t\sqrt{\pi}},
\end{equation*}
where the matrix $\Phib \in \R^{T \times T}$ is such that $[\Phib]_{ij}=\phi_{(\sigma^2+\gamma^2_i+\sigma^2+\gamma^2_j)^{1/2}}(\mu_i-\mu_j)$, and the $i$-th entry of vector $\thetab$ is $\phi_{(\sigma^2+\gamma^2_i+\gamma_t^2)^{1/2}}(\mu_i-\mu_t)$. By the same token, we can write the variance term as
\begin{equation*}
    IV(\hat{h}_t(x)) = \alphab^\top \Db \alphab,
\end{equation*}
where $\Db$ is a diagonal matrix with $[\Db]_{ii}=\frac{1}{n_i2\sqrt{\pi}}(\frac{1}{\sigma} - \frac{1}{\sqrt{\sigma^2+\gamma^2_i}})$. This completes the proof of Theorem \ref{the: mise}.

\subsubsection{Acknowledgements} The authors gratefully acknowledge the support of NSF Award \#2038625 as part of the NSF/DHS/DOT/NIH/USDA-NIFA Cyber-Physical Systems Program.

\bibliographystyle{splncs04}
\bibliography{mybibliography}

\begin{thebibliography}{10}
\providecommand{\url}[1]{\texttt{#1}}
\providecommand{\urlprefix}{URL }
\providecommand{\doi}[1]{https://doi.org/#1}

\bibitem{amiri2018density}
Amiri, A., Dabo-Niang, S.: Density estimation over spatio-temporal data
  streams. Econometrics and Statistics  \textbf{5},  148--170 (2018)

\bibitem{boedihardjo2008framework}
Boedihardjo, A.P., Lu, C.T., Chen, F.: A framework for estimating complex
  probability density structures in data streams. In: Proceedings of the 17th
  ACM Conference on Information and Knowledge Management. pp. 619--628 (2008)

\bibitem{UCRArchive2018}
Dau, H.A., Keogh, E., Kamgar, K., Yeh, C.C.M., Zhu, Y., Gharghabi, S.,
  Ratanamahatana, C.A., Chen, Y., Hu, B., Begum, N., Bagnall, A., Mueen, A.,
  Batista, G., Hexagon-ML: The {UCR} time series classification archive
  (October 2018), \url{https://www.cs.ucr.edu/~eamonn/time_series_data_2018/}

\bibitem{devlin2018bert}
Devlin, J., Chang, M.W., Lee, K., Toutanova, K.: Bert: Pre-training of deep
  bidirectional transformers for language understanding. arXiv preprint
  arXiv:1810.04805  (2018)

\bibitem{heinz2008cluster}
Heinz, C., Seeger, B.: Cluster kernels: Resource-aware kernel density
  estimators over streaming data. IEEE Transactions on Knowledge and Data
  Engineering  \textbf{20}(7),  880--893 (2008)

\bibitem{marron1992exact}
Marron, J.S., Wand, M.P.: Exact mean integrated squared error. The Annals of
  Statistics  \textbf{20}(2),  712--736 (1992)

\bibitem{qahtan2016kde}
Qahtan, A., Wang, S., Zhang, X.: {KDE}-track: An efficient dynamic density
  estimator for data streams. IEEE Transactions on Knowledge and Data
  Engineering  \textbf{29}(3),  642--655 (2016)

\bibitem{vaswani2017attention}
Vaswani, A., Shazeer, N., Parmar, N., Uszkoreit, J., Jones, L., Gomez, A.N.,
  Kaiser, {\L}., Polosukhin, I.: Attention is all you need. Advances in neural
  information processing systems  \textbf{30} (2017)

\bibitem{velivckovic2017graph}
Veli{\v{c}}kovi{\'c}, P., Cucurull, G., Casanova, A., Romero, A., Lio, P.,
  Bengio, Y.: Graph attention networks. arXiv preprint arXiv:1710.10903  (2017)

\bibitem{wand1994kernel}
Wand, M.P., Jones, M.C.: Kernel Smoothing. CRC press (1994)

\bibitem{wang2022takde}
Wang, Y., Ding, Y., Shahrampour, S.: Takde: Temporal adaptive kernel density
  estimator for real-time dynamic density estimation. arXiv preprint
  arXiv:2203.08317  (2022)

\bibitem{zhou2003m}
Zhou, A., Cai, Z., Wei, L., Qian, W.: M-kernel merging: Towards density
  estimation over data streams. In: Proceedings of the Eighth International
  Conference on Database Systems for Advanced Applications, 2003.(DASFAA 2003).
  pp. 285--292 (2003)

\end{thebibliography}

\end{document}